\pdfoutput=1

\documentclass[11pt]{article}

\usepackage{ACL2023}

\usepackage{times}
\usepackage{latexsym}
\usepackage{booktabs}
\usepackage{amsmath}
\usepackage{amssymb}
\usepackage{graphicx}

\usepackage[T1]{fontenc}

\usepackage[utf8]{inputenc}

\usepackage{microtype}

\usepackage{inconsolata}

\title{Contextual word2vec}
\title{Backpack-of-Words Networks}
\title{Backpack-of-Words}
\title{Backpack Representations: Orderly, Weighted Bag-of-Words}
\title{Language Modeling with Backpack Representations}
\title{\modelname Language Models}

\usepackage{xspace}

\newcommand{\modelname}{Backpack\xspace}

\newcommand{\V}{\mathcal{V}}
\newcommand{\x}{\mathbf{x}}
\newcommand{\h}{\mathbf{h}}
\renewcommand{\v}{\mathbf{v}}
\renewcommand{\o}{\mathbf{o}}

\newcommand\dash{---}

\newcommand{\AnD}{\hskip 0.9em plus 0.15fil minus 0.25em}
 \author{John Hewitt \AnD John Thickstun \AnD Christopher D. Manning \AnD Percy Liang \\
 Department of Computer Science, Stanford University \\
 \{\texttt{johnhew,jthickstun,manning,pliang}\}\texttt{@cs.stanford.edu}}

\begin{document}
\maketitle
\begin{abstract}

We present \textit{\modelname{s}}: a new neural architecture that marries strong modeling performance with an interface for interpretability and control.
\modelname{s} learn multiple non-contextual \textit{sense} vectors for each word in a vocabulary,
and represent a word in a sequence as a context-dependent, non-negative linear combination of sense vectors in this sequence.
We find that, after training, sense vectors specialize, each encoding a different aspect of a word.
We can interpret a sense vector by inspecting its (non-contextual, linear) projection onto the output space, 
and intervene on these interpretable hooks to change the model's behavior in predictable ways. %
We train a 170M-parameter \modelname language model on OpenWebText, matching the loss of a GPT-2 small (124M-parameter) Transformer. %
On lexical similarity evaluations, we find that \modelname sense vectors outperform even a 6B-parameter Transformer LM's word embeddings.
Finally, we present simple algorithms that intervene on sense vectors to perform controllable text generation and debiasing. %
For example, we can edit the  sense vocabulary to tend more towards a topic, or localize a source of gender bias to a sense vector and globally suppress that sense.

\end{abstract}

\section{Introduction}

Consider the prefix \textit{The CEO believes that \_\_\_}, and the problem of debiasing 
 a neural language model's distribution over \textit{he/she}. 
Intuitively, the bias for \textit{he} originates in the word \textit{CEO}, because replacing \textit{CEO} with \textit{nurse} flips the observed bias.
A successful intervention to debias \textit{CEO} must reliably apply in all contexts in which the word \textit{CEO} appears; ideally we would want to make a \textbf{non-contextual} change to the model that has predictable effects in \textbf{all contexts}.
In general, 
in all aspects of interpretability and control, it is desirable to make interventions with a tractable interface (e.g., non-contextual representations) that apply globally.

\begin{figure}
\includegraphics[width=\linewidth]{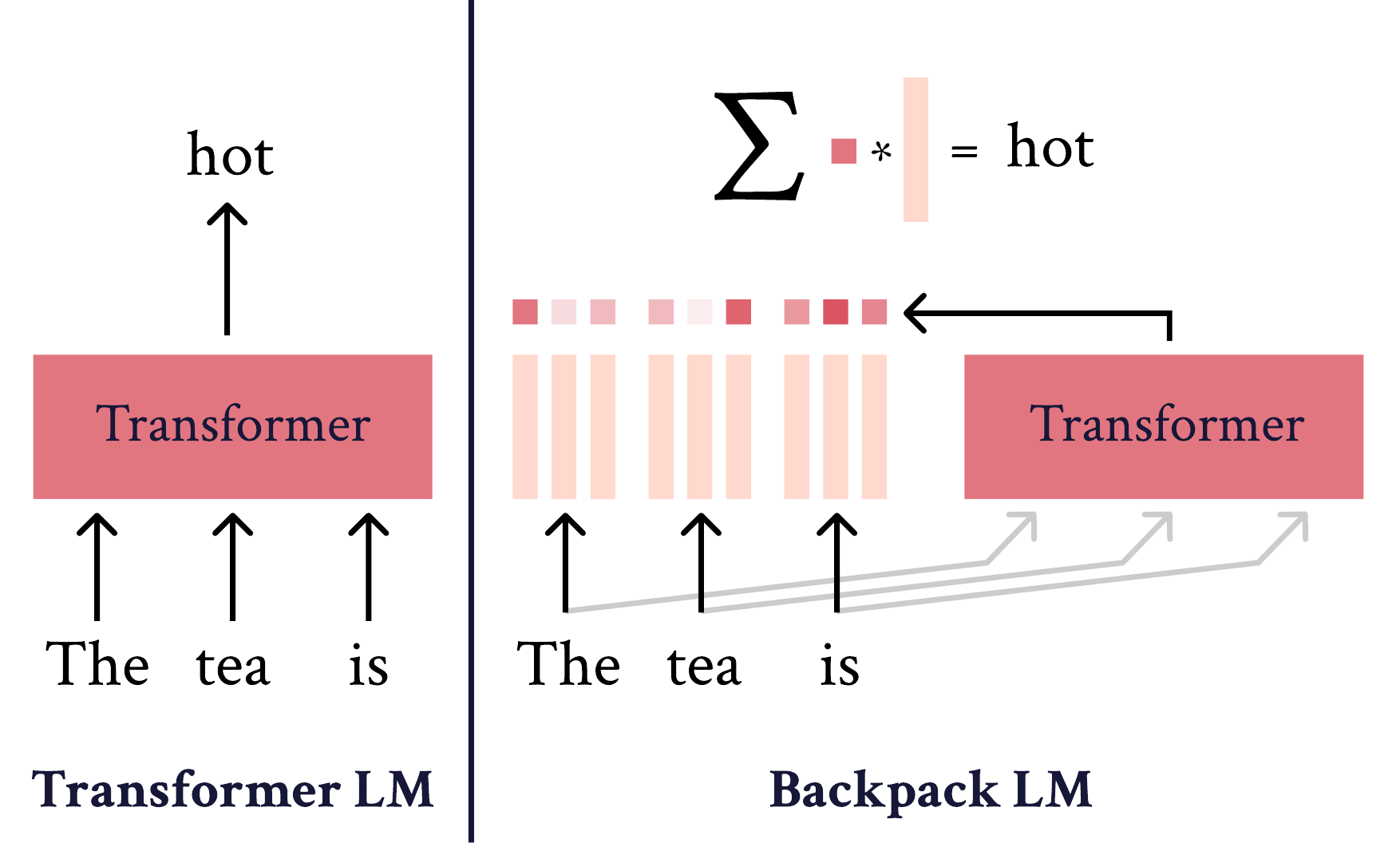}
\caption{\label{fig1}Transformers are monolithic functions of sequences. In \modelname{s}, the output is a weighted sum of non-contextual, learned word aspects.}
\vspace{-.9mm}
\end{figure}

Such interventions are difficult in Transformer models \cite{vaswani2017attention} because their contextual representations are monolithic functions of their input.
 Almost any intervention on the model has complex, non-linear effects that depend on context.
 We would instead like models that enable precise, rich interventions that apply predictably in all contexts, and are still expressive, so they are a viable alternative to  Transformers.

\begin{table*}

\small
\centering
\begin{tabular}{c c c c c}
\toprule
\multicolumn{5}{c}{A few senses of the word \textit{science}} \\
\midrule
Sense 3&                Sense 7&              Sense 9&              Sense 10&           Sense 8       \\
\midrule
fiction&          replication&    religion&       settled&       clones      \\
fictional&        citation&       rology&         sett&         experiments\\ 
Fiction&          Hubble&         hydra&          settle&       mage \\   
literacy&         reprodu&        religions&      unsett&       experiment \\ 
denial&           Discovery&      nec&            Sett&         rats \\

\midrule
\end{tabular}\hspace{20pt}%
\begin{tabular}{p{6.2cm}}
\toprule
\textit{\textbf{MacBook$_{\text{HP}}$}} = \textit{\textbf{MacBook}} $-$ \textit{\textbf{Apple}} $+$ \textit{\textbf{HP}}\\
\midrule
\textbf{The MacBook is best known for} its form factor, but HP has continued with its Linux-based computing strategy. HP introduced the Hyper 212 in 2014 and has continued to push soon-to-be-released 32-inch machines with Intel's Skylake processors.\\
\midrule
\end{tabular}
\caption{\label{table_taste_fig2}Examples of the rich specialization of sense vectors representing the word \textit{science}, and an example of editing sense vectors non-contextually (changing MacBook to be associated with HP) and having the resulting \textit{contextual} predictions change.}
\end{table*}

We address these challenges with a new neural architecture, the \textit{\modelname}, for which predictions are log-linear combinations of non-contextual representations.
We represent each word in a vocabulary as a set of non-contextual \textit{sense vectors} that represent distinct learned aspects of the word.
For example, sense vectors for the word ``science'' could encode types of science, connections to technology, notions of science being ``settled,'' or different aspects of the scientific process (replication or experiment) (Table~\ref{table_taste_fig2}).
Sense vectors do not learn classic word sense, but more general aspects of a word's potential roles in different contexts; in fact, they 
can be seen as a multi-vector generalization of classic word vectors \cite{mikolov2013efficient}.\footnote{Our code, sense vectors, language model weights, and demos are available at \url{https://backpackmodels.science}.} %

To make interventions on sense vectors behave predictably in different contexts, a \modelname represents each word in a sequence as a \textbf{linear combination} of the sense vectors for all words in the sequence.
The expressivity of a \modelname comes from the network that computes the weights of the linear combination as a function of the whole sequence; for example, in all our experiments we use a Transformer for this. %
Since sense vectors are softly selected depending on the context, they can specialize; each sense can learn to be predictively useful in only some contexts.
The log-linear contribution of senses to  predictions then implies that the interventions on sense vectors we demonstrate in Section~\ref{sec:control} apply identically (up to a non-negative scalar weight) regardless of context.

Our experiments demonstrate the expressivity of \modelname language models, and the promise of interventions on sense vectors for interpretability and control.
In Section~\ref{sec:training} we train \modelname language models on 50B tokens (5 epochs) of OpenWebText; a \modelname with 124M parameters in the contextual network (and 46M parameters for sense vectors) achieves the perplexity of a 124M-parameter Transformer; thus one pays for more interpretability with a larger model size. %
In Section~\ref{sec:sense-structure}, we show that sense vectors specialize to encode rich notions of word meaning.
Quantitatively, on four lexical similarity datasets (e.g., SimLex999), sense vectors of a 170M parameter \modelname outperform word embeddings of the 6B-parameter GPT-J-6B Transformer, and approach the performance of state-of-the-art specialized methods for this task.
Finally, in Section \ref{sec:control} we show that sense vectors offer a control mechanism for \modelname language models.
For example, stereotypically gendered profession words (e.g., ``CEO'' or ``nurse'') tend to learn a sense vector associated with this gender bias; by downscaling this sense vector, we greatly reduce disparity in contextual predictions in a limited setting.

\section{The \modelname Architecture} \label{sec_modelname}

In this section, we define the general form of the \modelname architecture. %
We then show how continuous bag-of-words word2vec (CBOW) \citep{mikolov2013efficient} and Self-Attention-Only networks \cite{elhage2021mathematical,olsson2022context} are special cases of \modelname{s}.

\subsection{\modelname General Form}\label{sec:backpack-form}
A \modelname is a parametric function that maps a sequence of symbols $\x_{1:n} =\left(\x_1,\dots, \x_n\right)$ to a sequence of vectors $\o_{1:n}=\left(\o_1,\dots, \o_n\right)$, where each symbol $\x_i$ belongs to a finite vocabulary $\V$ and $\o_i\in\mathbb{R}^d$.
We call $\o_i$ the \emph{\modelname representation} of $\x_i$ in the context of a sequence $\x_{1:n}$.

\paragraph{Sense vectors.}
For each $\x \in \V$, a \modelname{} constructs $k$ \textit{sense} vectors
\begin{align}\label{eqn:content}
C(\x)_1,\dots, C(\x)_k,
\end{align}
where $C : \V \to \mathbb{R}^{k\times d}$. %
Sense vectors are a multi-vector analog to classic non-contextual word representations like word2vec or GloVe: we make this analogy precise in Section~\ref{sec:cbow}.

\paragraph{Weighted sum.}
For a sequence $\x_{1:n}$, the representation $\o_i$ of element $\x_i$ is a weighted sum of the predictive sense vectors for the words in its context: given \emph{contextualization weights} $\alpha \in \mathbb{R}^{k\times n\times n}$,
\begin{align}
\o_i = \sum_{j=1}^{n}\sum_{\ell=1}^k \alpha_{\ell ij} C(\x_j)_\ell.
\end{align}
The contextualization weights $\alpha_{\ell ij}$ of a \modelname are themselves defined by a (non-linear) \textit{contextualization function} of the entire sequence $\x_{1:n}$: %
\begin{align}\label{eqn:contextualization}
\alpha = A(\x_{1:n}),
\end{align}
where $A : \V^n \to \mathbb{R}^{k\times n\times n}$.

The name ``\modelname'' is inspired by the fact that a backpack is like a bag---but more orderly.
Like a bag-of-words, a \modelname representation is a sum of non-contextual senses; but a \modelname is more orderly, because the weights in this sum depend on the ordered sequence.

\paragraph{\modelname Models.}
A \emph{\modelname model} is a probabilistic model that defines probabilities over some output space $\mathcal{Y}$ as a log-linear function of a \modelname representation $\o_{1:n} \in \mathbb{R}^{n\times d}$:
\begin{align}
p(\mathbf{y}|\o_{1:n}) = \text{softmax}\left(E( \o_{1:n})\right),
\end{align}
 where $\mathbf{y} \in \mathcal{Y}$ and $E : \mathbb{R}^{n\times d} \to \mathbb{R}^{|\mathcal{Y}|}$ is a linear transformation.
Because \modelname models are log-linear in their representations, the sense vectors contribute log-linearly to predictions. This allows us to inspect a sense vector by projecting it onto the vocabulary via $E$ and observe exactly how it will contribute to predictions in any context.

Models parameterized by the prevailing deep neural architectures\dash including LSTMs \cite{hochreiter1997long} and Transformers\dash are not \modelname{s} because their output representations are (relatively) unconstrained functions of the entire sequence.
By contrast, \modelname models may seem limited in expressivity: the representations $\o_i$ are scalar-weighted sums of non-contextual vectors $C(\x_j)_\ell$. Contextual relationships between sequence elements can only be expressed through the weights $\alpha=A(\x_{1:n})$.
Nevertheless, our experiments show that an expressive contextualization weight network can represent complex functions by weighted sums of sense vectors, e.g., our 170M parameter \modelname LM uses a 124M-parameter Transformer to compute $\alpha$, and achieves the loss of a 124M-parameter Transformer LM.

To place \modelname{s} in some historical context, we now show how two existing architectures can be described as \modelname{s}. 

\subsection{Continuous Bag-of-Words is a \modelname}\label{sec:cbow}
The continuous bag-of-words word2vec model defines a probability distribution over a center word $\x_c \in \V$ conditioned on $n$ context words $\x_{1:n}$.\footnote{Context in this setting is usually defined as words surrounding the center word.}
The model proceeds to (1) construct vector embeddings $\v_\x$ for each $\x\in\V$, and (2) uniformly average the embeddings of the context words to predict the center word:
\begin{align}
&\overline{\v}_{\x_c} = \sum_{i=1}^{n} \frac{1}{n}\v_{\x_i},\\
  &p(\x_c \mid \x_{1:n}) = \text{softmax}(U\overline{\v}_{\x_c}),
\end{align}
where $U\in\mathbb{R}^{\V\times d}$.
We see that $\overline{\v}_{\x_c}$ is a \modelname representation by setting $C(\x) = \v_\x \in\mathbb{R}^{1\times d}$ in Equation \eqref{eqn:content} using a single sense vector ($k=1$) and setting the contextualization weights in Equation \eqref{eqn:contextualization} to be uniform: $\alpha_{\ell ij}=\frac{1}{n}$.

This connection to CBoW foreshadows the emergence of linguistic structures in the predictive sense vectors of \modelname models, just as these structures emerge in CBoW \citep{mikolov2013efficient}.

\subsection{Single-Layer Self-Attention is a \modelname}
The \modelname structure---define sense vectors (values), and use the sequence to determine how to sum them (weights)---may remind the reader of a single layer of self-attention.
The key-query-value self-attention function is as follows:
\begin{align}
&\o_j = \sum_{i=1}^{n}\sum_{\ell=1}^{k} \alpha_{\ell ij} O V^{(\ell)} \x_j\\
&\alpha_\ell = \text{softmax}(\x^\top K^{(\ell)\top} Q^{(\ell)} \x),\label{eqn:attn}
\end{align}
where $\x\in\mathbb{R}^{n\times d}$  is (overloaded) to be a non-contextual embedding of the sequence, $O\in\mathbb{R}^{d\times d/k}$, and $V^{(\ell)}\in \mathbb{R}^{d/k \times d}$, where $k$ is the number of attention heads.
The self-attention function is a \modelname with $C(\x_j)_\ell= OV^{(\ell)} \x_j$.
Self-attention-only networks are studied in the context of, e.g., mechanistic interpretability \cite{elhage2021mathematical}.
A Transformer composes blocks of self-attention and non-linear feed-forward layers that combine information from the whole sequence; unlike a Transformer, the contextualization weights of a \modelname each select a non-contextual sense of a single word.

\section{Language Modeling with \modelname{s}}\label{sec:lm}

In this section, we define a neural autoregressive language model parameterized by a \modelname.
We use the standard softmax parameterization of the probability over the next token in a sequence, with a weight matrix $E\in\mathbb{R}^{d\times |\V|}$ that maps a representation $\o_j \in \mathbb{R}^d$ to logits $E^\top \o_j \in \mathbb{R}^{|\V|}$:
\begin{align}\label{eqn:backpack-lm}
p(\x_j \mid \x_{1:j-1}) = \text{softmax}(E^\top \o_j).
\end{align}
Recall (Section \ref{sec:backpack-form}) that \modelname representations $\o_j$ are defined by sense vectors $C(\x)$ and contextualization weights $\alpha_j$. In Section \ref{sec:param-sense} we describe a parameterization of $C$ for the predictive sense vectors in Equation~\eqref{eqn:content}, and in Section \ref{sec:param-weights} we describe a parameterization of $A$ for the contextualization weight network in Equation~\eqref{eqn:contextualization}.
When $\o_j$ is parameterized by a \modelname, we call a model of the form given by Equation~\eqref{eqn:backpack-lm} a \emph{\modelname LM}.

\subsection{Parameterizing senses}\label{sec:param-sense}
For the sense function $C : \mathcal{V} \to \mathbb{R}^{k\times d}$, we embed each $\x\in\V$ into $\mathbb{R}^d$ and pass these embeddings though a feed-forward network $\text{FF} : \mathbb{R}^d \to \mathbb{R}^{k\times d}$: %
\begin{align}
C(\x) = \text{FF}(E\x),
\end{align}
where the embedding/projection matrix $E$ is tied to the output matrix in Equation \eqref{eqn:backpack-lm} \citep{press2017using}.
Note that we could define all $k\times |\V|$ sense vectors using a lookup table, but this would be an enormous number of parameters as $k$ grows large.
Instead, we embed the words as $E\x \in \mathbb{R}^d$, and then blow them up to $\mathbb{R}^{d\times k}$ using shared weights.
This may explain the related sense roles observed for different word types in Section~\ref{sec_visualizing_senses}.

\subsection{Parameterizing contextualization weights}\label{sec:param-weights}

We parameterize $A  : \V^n \to \mathbb{R}^{k\times n\times n}$ using a standard Transformer, followed by a layer of multi-headed key-query self-attention. %
That is, we pass an embedded sequence through a Transformer
\begin{align}
    \h_{1:n} = \text{Transformer}(E\x_{1:n}) \label{eqn_contextual_hiddens}
\end{align}
(with proper autoregressive masking and some position representation) and compute $A(\x_{1:n}) = \alpha$, where 
\begin{align}
    &\alpha_{\ell} = \text{softmax}(\h_{1:n} K^{(\ell)\top} Q^{(\ell)} \h_{1:n}^\top), \label{eqn_weight_self_attention}
\end{align}
for each predictive sense $\ell=1,\dots,k$ with matrices $K^{(\ell)},Q^{(\ell)}\in\mathbb{R}^{d\times d/k}$.
We can think of the $k$ senses as heads and, for each head, the contextualization weights define a distribution of attention over words.\footnote{Note that the sense weights are normalized (1) independently for each sense, and (2) to sum to one over the sequence length.}

\section{Experiments Training \modelname LMs}\label{sec:training}

In this section we specify the hyperparameters used to train \modelname and Transformer language models  (Section~\ref{sec_expts_models}), data and optimization procedure (Section~\ref{sec_expts_data_opt}), evaluations (Section~\ref{sec_expts_eval}) and results (Section~\ref{sec_expts_results}).
We also show the necessity of learning $k>1$ sense vectors to achieve strong language modeling performance (Section~\ref{sec_ablation_results}).

\begin{table*}[t!]
\small
\centering
\begin{tabular}{l c c c c c}
\toprule
Model & OpenWebText PPL $\downarrow$ & LAMBADA PPL $\downarrow$ & LAMBADA ACC $\uparrow$ & Wikitext PPL $\downarrow$ & BLiMP $\uparrow$\\
\midrule
\modelname-Micro & \bf 31.5 
 & \bf 110 & \bf24.7 & \bf 71.5 & 75.6\\
Transformer-Micro & 34.4 & 201& 21.3 & 79.5 & \bf 77.8\\
\midrule
\modelname-Mini & \bf 23.5 & \bf 42.7 & \bf 31.6 & \bf 49.0 & 76.2 \\
Transformer-Mini  & 24.5 & 58.8 & 29.7 & 52.8 & \bf 80.4\\
\midrule
\modelname-Small  & \bf 20.1 & \bf 26.5 & \bf 37.5 & \bf 40.9 & 76.3 \\
Transformer-Small  & \bf 20.2 & 32.7 & 34.9 & 42.2 & \bf 81.9\\
\bottomrule
\end{tabular}
\caption{\label{table_main_results} Language modeling performance; all models trained for 100k steps, 500K token batch size, on OWT. For PPL, lower is better; for accuracy, higher is better. Note that models are not parameter-comparable; each Backpack has a matched-size Transformer in its contextualization network.}
\end{table*}

\subsection{Models} \label{sec_expts_models}
We train three Transformer baseline models, which we label Micro (30M parameters), Mini (70M parameters), and Small (124M parameters; the same size as GPT-2 small).
We also train Micro (40M), Mini (100M), and Small (170M) \modelname language models, for which the weighting function (Equation~\ref{eqn_contextual_hiddens}) is parameterized using the corresponding Transformer, and almost all extra parameters are in the non-contextual sense vectors.\footnote{There are a negligible number of additional parameters in the final key-query \modelname operation (Equation~\ref{eqn_weight_self_attention})).}
\modelname{s} thus cost extra parameters and compute beyond their underlying contextualization network.
Except where stated, we use $k=16$ sense vectors in all \modelname{s} (Section~\ref{appendix_training_details}).

We use a reduced sequence length of 512 for all models, and the 50,257-subword GPT-2 tokenizer.
Model hidden dimensionalities, layer counts, and head counts are reported in Table~\ref{table_model_hyperparams}.

\subsection{Data \& Optimization} \label{sec_expts_data_opt}
We train all models on OpenWebText \cite{Gokaslan2019OpenWeb}, a publicly available approximate reconstruction of the English WebText corpus used to train the GPT-2 family of models \cite{radford2019language}.
We use a batch size of 524,288 tokens, and train all models for 100,000 gradient steps for a total of 52B tokens; training for longer is known to make marginal difference for small models \cite{hoffmann2022training}.
The size of OpenWebText means this is roughly 5 epochs.
We use cross-entropy loss and the AdamW optimizer, with a warmup of 5,000 steps and linear decay to zero.

\subsection{Evaluations} \label{sec_expts_eval}
Before our experiments in interpretability and control, we check the expressivity of \modelname{s}.
We evaluate models on perplexity for a held out set of OpenWebText, perplexity and accuracy for the (OpenAI variant of) LAMBADA evaluation of long-distance dependencies \cite{radford2019language,paperno2016lambada}, perplexity on Wikitext \cite{merity2017pointer}, and BLiMP English linguistic competence accuracy \cite{warstadt2020blimp} evaluated using the EleutherAI harness \cite{gao2021framework} (Version 1).

\subsection{Discussion} \label{sec_expts_results}

Comparing each \modelname LM to a Transformer LM of equivalent specification to the \modelname's contextualization network, we see that the \modelname performs roughly as well (Table~\ref{table_main_results}).
Again, the \modelname has more parameters, a tax for the interface provided by sense vectors.
During training, we find that \modelname language models take longer to converge than Transformers.
Curiously, while the Small \modelname  and Transformer achieve almost identical OWT perplexity, the \modelname language models perform substantially better on LAMBADA and Wikitext, but worse on BLiMP.

\subsection{Effect of varying the number of senses} \label{sec_ablation_results}
To study the impact of the number of sense vectors on language modeling performance, we train Mini-sized \modelname language models on a reduced schedule of 50,000 gradient steps, for $k\in\{1,4,16,64\}$ sense vectors.
The perplexities for $k=1,4, 16,64$ are 38.6, 29.3, 26.0, and 24.1, demonstrating the necessity of a non-singleton set of sense vectors.
 Table~\ref{table_p_sense_vecs_results} contains the full results.

\section{Emergent Structure in Sense Vectors}\label{sec:sense-structure}

\begin{table*}
\small
\centering
\begin{tabular}{c c c c c c c c c}
\multicolumn{4}{c}{Sense 12 (\textit{relatedness})} & & \multicolumn{4}{c}{Sense 14 (\textit{Verb objects, nmod nouns})}\\ 
\cmidrule{1-4}
\cmidrule{6-9}
\textit{tasty} & \textit{quickly} & \textit{Apple} & \textit{believe} & & \textit{build} & \textit{attest} & \textit{importance}  & \textit{appreciate}\\ 
\cmidrule{1-4}
\cmidrule{6-9}
tasty&        quick&      Apple&        belief&     & bridges&       worthiness&       maintaining& finer \\
culinary&     quickest&   Apple&        Belief&     & wall&          Published&        wellbeing&   nuance \\
tasted&       quick&      iPhone&       beliefs&    & lasting&       superiority&      teamwork&    beauty \\
delicious&    quicker&    iPhone&       believing&  & ig&            accuracy&         plurality&   irony \\
taste&        fast&       iPhones&      believe&    & rapport&       validity&         upholding&   simplicity \\
\cmidrule{1-4}
\cmidrule{6-9}
\end{tabular}

\vspace{10pt}

\begin{tabular}{c c c c c c c c c c}
\multicolumn{3}{c}{Sense 3 (\textit{next wordpiece})} & & \multicolumn{3}{c}{Sense 7 (\textit{Proper Noun Associations})}\\ 
\cmidrule{1-3}
\cmidrule{5-7}
\textit{pizza}& \textit{interest}& \textit{the}&  & \textit{Apple}&    \textit{Obama}&  \textit{Messi} &     \\
\cmidrule{1-3}
\cmidrule{5-7}
cutter&         rate&     slightest&       &      macOS&     Dreams&   Messi&                               \\
tracker&        rates&    same&            &      iCloud&    Barack&   Argentina&                              \\
iol&            groups&   entirety&        &      Siri&      Ob&       Mess&                              \\
makers&         waivers&  rest&            &      iOS&       Michelle&  Barcelona&                              \\
maker&          waiver&   latter&          &      tv&        Jeremiah&  iesta&                              \\
\cmidrule{1-3}
\cmidrule{5-7}
\end{tabular}
\caption{\label{table_sense_visualization}Visualization of how the same sense index across many words encodes fine-grained notions of meaning, relatedness, and predictive utility.
Each sense is given a label thought up by the authors, and for a few words, the target words that are highest scored by the sense vector.}
\end{table*}

\begin{table}
\centering
\small
\begin{tabular}{l c c c c}
\toprule
Model & SL999 & SV3500 & RG65 & WS353\\
\midrule
\multicolumn{5}{l}{\textit{Classic Non-Contextual Embeddings}}\\
word2vec &0.442&0.367&0.679&0.684\\ 
 GloVe&0.371&0.227& 0.687&0.607\\
\midrule
\multicolumn{5}{l}{\textit{Embeddings from large existing models}}\\
 GPT2-1.5B&0.523&0.418&0.670&0.706\\ 
 GPT-J-6B&0.492&0.374& \bf 0.766&0.673\\
 \midrule
\multicolumn{5}{l}{\textit{Embeddings from our models + baseline Transformer}}\\
 Trnsf 124M &0.478&0.363&0.634&0.681\\
Sim$_{12}$ (ours) &0.522&0.471& 0.754&\bf 0.749\\
Sim$_{14}$ (ours) &0.500&\bf 0.502&0.591&0.655\\
Sim$_{\text{min}}$  (ours) &\bf 0.540&0.471&0.653&0.607\\
\midrule[\heavyrulewidth]
\multicolumn{5}{l}{\textit{Special-purpose SOTA models}}\\
SOTA (Single) &  0.554& 0.473 &0.835 & 0.764\\
SOTA (Multi) &  0.605 & 0.528 & - & 0.807\\
\bottomrule
\end{tabular}
\caption{\label{table_simlex_results}Results on lexical similarity evaluation. All numbers are Spearman correlations; higher is better.}
\end{table}

\modelname language model sense vectors are not trained using a supervised notion of word sense, but implicitly specialize to encode different shades of a word's predictive use.
In this section, we qualitatively examine sense vectors (Section~\ref{sec_visualizing_senses}) and quantitatively demonstrate their effectiveness in computing lexical similarity and relatedness (Section~\ref{sec_lexical_evals}). 
Taken together, this suggests that sense vectors can provide a high-level interface for intervention, which we explore in Section~\ref{sec:control}. %

\subsection{Visualizing Senses} \label{sec_visualizing_senses}

Empirically, trained \modelname models associate specific sense vector indices with different roles for prediction.
We interpret these roles by picking a sense $\ell$ of a word $\x$, and projecting this sense onto the word embeddings: $E^\top C(\x)_\ell \in \mathbb{R}^{|\V|}$. Note that this is \textit{exactly} (up to a scalar) how this sense contributes to any prediction of the model.
We interpret a sense vector's role by reporting the words with the highest score under this projection.

Table~\ref{table_sense_visualization} visualizes a few of these senses. %
For example, sense 12 seems to encode a broad notion of relatedness for almost all words; sense 3 encodes particulars of the bigram distribution given $\x$; sense 14 seems to encode both associated objects for verbs, and noun modifier dependency children for nouns.
In Section~\ref{sec_lexical_evals} we show that sense 14 encodes a powerful notion of verb similarity.

\subsection{Lexical Relationship Tests} \label{sec_lexical_evals}

Classic lexical-relatedness and similarity tests measure the extent to which a similarity function on pairs of words correlates with human-elicitied notions of similarity.
Similarity functions derived from word embeddings are evaluated by Spearman correlation between the predicted and true similarity rank-order.
Early non-contextual embeddings like COALS \cite{rohde2006improved}, word2vec \cite{mikolov2013efficient}, and GloVe \cite{pennington2014glove} have recently been  outperformed by word embeddings derived by distillation of contextual networks \cite{bommasani2020interpreting,gupta2021obtaining,chronis2020bishop}. 
We evaluate \modelname LM sense vectors on similarity datasets SimLex999 \cite{hill2015simlex}, SimVerb3500 \cite{gerz2016simverb}, and relatedness datasets RG65 \cite{rubenstein1965contextual} and \cite{agirre2009study}.

\paragraph{Sense$_\ell$ Cosine.} For all $\ell\in\{1,\dots,k\}$, we define a similarity function based only on sense $\ell$:
\begin{align}
\text{Sim}_\ell(\x, \x') = \text{cossim}(C(\x)_\ell,  C(\x')_\ell),
\end{align}
where cossim is cosine similarity.
Intuitively, we expect that some senses may specialize to learn lexical relatedness or similarity.

\paragraph{Minimum Sense Cosine.}
Because each sense encodes a different aspect of a word's meaning, we might expect that highly similar words are similar across \textit{all} senses.
We test for this strong form of similarity using
\begin{align}
\text{Sim}_{\text{min}}(\x,\x') =\min_{\ell} \text{Sim}_\ell(\x,\x')
\end{align}

\paragraph{Other methods.}
We evaluate embeddings from the tied softmax/embedding matrices of the much larger GPT-2-1.5B \cite{radford2019language} and GPT-J-6B \cite{wang2021gptj}, classic word embeddings (from \citet{bommasani2020interpreting}) and state-of-the art specialized methods using either a single vector per word \cite{gupta-2021-multilingual} or many vectors \cite{chronis2020bishop}.

\paragraph{Discussion.}
Sense $12$ (the ``synonym'' sense) performs well across datasets, matching or outperforming embeddings like GPT-2-1.5B and GPT-J-6B (Except GPT-J-6B on RG-65).
Sense $14$, the ``verb objects'' sense, performs best on just verb similarity (VerbSim3500), and the minimum similarity over senses works especially well on noun lexical similarity (SimLex999.)
Our methods approach the performance of state-of-the-art methods; despite being trained for a very different task, sense vectors encode substantial lexical information (Table~\ref{table_simlex_results}).

\section{Sense Vectors for Control}\label{sec:control}

In this section, we demonstrate several proof-of-concept methods that leverage sense vectors for controlling LM behavior.

\subsection{Topic-controlled generation}

Given a bag-of-words target $b\in \mathbb{R}^{|\mathcal{V}|}$, e.g., \textit{arts, culture}, we would like to bias generation towards sequences related to concepts related to these terms.
Our algorithm proceeds in three parts.
First, we sort sense vectors by log-probability assigned to $b$, that is,  $b^\top(E^\top C(\x)_\ell)$.\footnote{We divide this term by the maximum absolute log-probability of the sense vector, $\max_{x\in\mathcal{V}} \x^\top(E^\top C(\x)_\ell)$.}
Second, based on the scores, we assign a re-weighting factor $\delta$ to each sense; senses with the higher scores weighted more. (See Section~\ref{appendix_generation_details} for details.)
Third, we generate from the \modelname using the re-weighted sense vectors, reducing $\delta$ back to $1$ as the topic is introduced.
The updated backpack equation is
\begin{align}
\o_i = \sum_{j=1}^{n}\sum_{\ell=1}^k \alpha_{\ell ij}\delta_{\ell ij} C(\x_j)_\ell,
\end{align}
where $\delta_{ij\ell}$ is the  re-weighting.
Intuitively, the semantic coherence of sense vectors may imply that upweighting senses with affinity to the target bag-of-words richly upweights related words and topics.
We give details as to how we perform the sense re-weighting and the annealing in Section~\ref{appendix_generation_details}.

\paragraph{Evaluation.} We use the label descriptors of the topic classifier of \citet{antypas2022twitter}, with 17 categories (\textit{sports, arts \& culture, health,\dots}), as the bag-of-words for control.
We evaluate control accuracy as the percent of generations to which the classifier assigns the correct topic label, and overall generation quality and diversity using MAUVE scores \citep{pillutla2021mauve}.\footnote{We concatenate generations across the 17 categories and compute MAUVE against OpenWebText validation examples.}

\begin{figure}
\includegraphics[width=\linewidth]{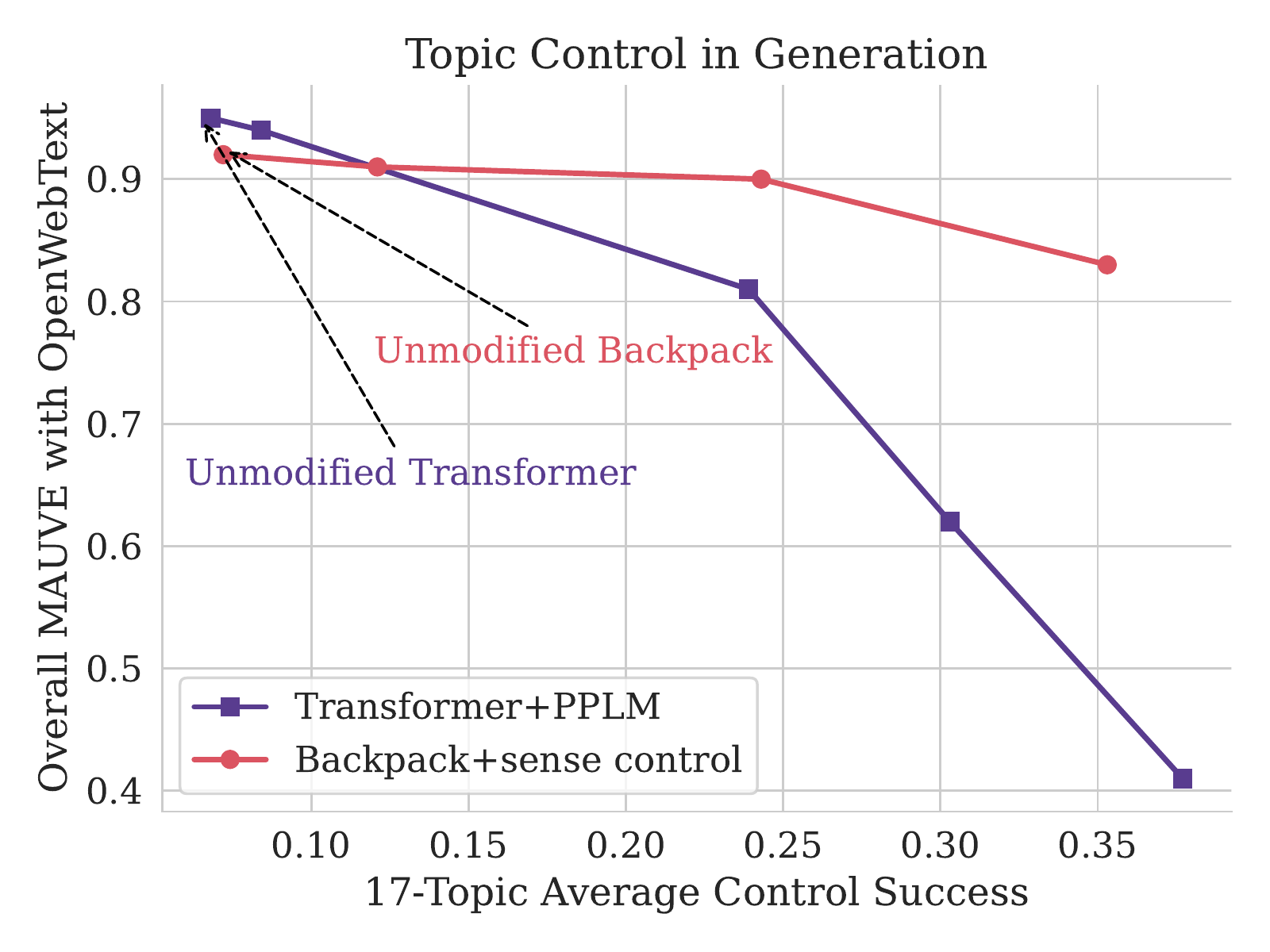}
\caption{\label{fig_topic} Results in controlling topic via sense intervention in Backpacks, and PPLM in Transformers. }
\end{figure}

\paragraph{Results.} We compare to Plug-and-Play Language Models (PPLM; \citet{dathathri2019plug}), a considerably slower, gradient-based control method using our Small Transformer model.
We generate 500 samples from each model for each topic across a range of strengths of control.
 We find that sense controlled generation provides at least as strong control as PPLM (Figure~\ref{fig_topic}), though the MAUVE scores of the unmodified Transformer are higher than the \modelname.)
Results and examples are provided in the Appendix in Tables~\ref{table_topic},~\ref{table_topic_ex1},~\ref{table_topic_ex2},~\ref{table_topic_ex3}.

\begin{figure*}
\includegraphics[width=\linewidth]{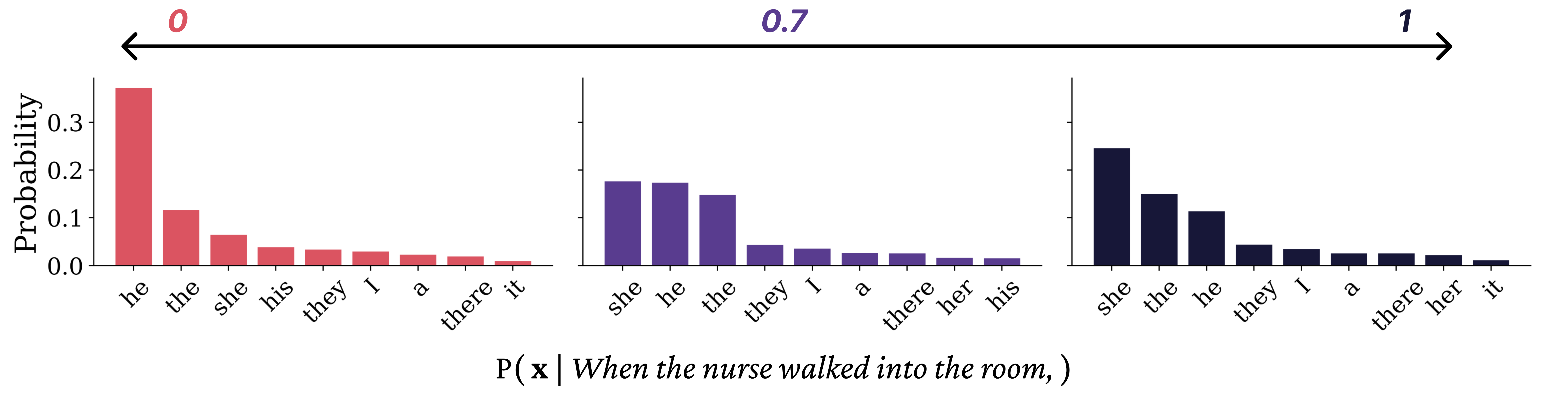}
\caption{\label{figure_gender_bias} The effect on the conditional probability distribution of a \modelname LM on the prefix \textit{when the nurse walked into the room,} of modulating the effect of sense 10 of \textit{ nurse} from 0 (totally removed) to 1 (original.)
}
\end{figure*}

\subsection{Mitigating gender bias}
Through inspection, we learned that sense vector 10 of many stereotypically gendered profession nouns (nurse, CEO, teacher) coherently express the stereotype through pronouns.
Table~\ref{table_gender_sense_example} gives examples of these senses.
We attempt to mitigate gender bias in \modelname behavior on these gendered profession nouns by \textit{turning down} sense 10 (multiplying by a scalar less than 1).

\begin{table}
\small
\centering
\begin{tabular}{l c c}
\toprule
Model & Bias Ratio $\downarrow$ & Reduction \%\\
\midrule
Unbiased & 1 & -\\
\midrule
\multicolumn{2}{l}{\textit{Transformer}}\\
Unmodified & 7.02 & - \\
Project-Nullspace & 6.72 & 5\%\\
Optimize-Nullspace & 7.02 & 0\%\\
\midrule
\multicolumn{2}{l}{\textit{Backpack}}\\
Unmodified & 4.34 & - \\
Remove-Sense10 & 2.88 & 44\%\\
Optimize-Sense10 & 2.16 & 65\%\\
\bottomrule
\end{tabular}
\caption{\label{table_gender} Pronoun-based gender bias reduction in a limited setting.}
\end{table}

We took an existing set of stereotypically gendered profession nouns from WinoBias \cite{zhao2018gender}, and constructed a simplified setting in which a single profession word is in each context, and a third-person nominative pronoun (e.g., he/she/they) is acceptable, e.g., \textit{My CEO said that\_\_}.
The full set of nouns and prompts is in Section~\ref{appendix_genderbias_details}.
We evaluate models on the average of the bias of probabilities of \textit{him} vs \textit{her} as follows:
\begin{align*}
\mathop{\mathbb{E}}
_{\x\in\text{prompts}}\left[ \max\left(\frac{p(\text{he}\mid \x)}{p(\text{she}\mid\x)}, \frac{p(\text{she}\mid \x)}{p(\text{he}\mid\x)}\right)\right].
\end{align*}

\paragraph{Baseline.} To debias a Transformer with an analogous method, we take inspiration from \citet{bolukbasi2016man}. We take $E\x_{\text{he}}-E\x_{\text{she}}$ as an estimate of a gender bias direction, and project the embedding $E\x_{\text{nurse}}$ either to the nullspace of this direction or only partially remove it.

\paragraph{Results.} A perfectly unbiased model would achieve ratio $1$, whereas the unmodified Transformer achieves $7$, and with nullspace projection, $6.72$ (Table~\ref{table_gender}).
Finding the optimal fraction of the gender bias direction to remove per profession does not improve further.
For \modelname{s}, we find that removing sense 10 from the profession word (setting it to zero) reduces the bias score from 4.34 to 2.88.
Learning the optimal removal fraction per profession achieves 2.16, for a total reduction of 65\%.\footnote{Curiously, \modelname{s} are overall less biased to begin with (in this setting); we don't have a strong hypothesis as to why.}
In Figure~\ref{figure_gender_bias}, we demonstrate the clear effect of ablating sense 10 on the most likely words in one of these contexts.\footnote{It is incidental that sense 10 encodes gender bias as opposed to another sense index; the consistency in index across words may be due to parameter sharing in $C$.}

\begin{table}
\centering
\small
\begin{tabular}{p{7.2cm}}
\toprule
\textbf{The MacBook is best known for} its form factor, but HP has continued with its Linux-based computing strategy. HP introduced the Hyper 212 in 2014 and has continued to push soon-to-be-released 32-inch machines with Intel's Skylake processors.\\
\midrule
\textbf{The MacBook didn't come into the picture until 2000}, when HP followed up with a 15-year flood of HP available laptops.\\
\midrule
\textbf{I was thinking about Brady's role on} the Colts before joining other high-profile signings. This is what McElhaney and I discussed.

McElhaney: Look, what I didn't mean by this is we didn't move. We think that we're getting a lot better, too. \\
\bottomrule
\end{tabular}
\caption{\label{table_macbook_control}Samples from a \modelname wherein \textit{Apple} has been projected out of the \textit{MacBook} sense embeddings, and replaced with \textit{HP}. Likewise with \textit{Brady}, \textit{Patriots}, and \textit{Colts}. Prompts are bolded.}
\end{table}

\subsection{Knowledge editing}
Sense vectors show promise for use in \textit{knowledge editing} \cite{de2021editing}---editing a model's predictions about world knowledge.
In particular, many associations with proper nouns can be localized to sense vectors in that noun.
In this qualitiative proof-of-concept, we edit the sense vectors of a target word $\x$ (e.g., \textit{MacBook} to remove associations with a word $\x_r$ (e.g., \textit{Apple}) and replace those associations with another word $\x_a$ (e.g., \textit{HP}).
Intuitively, this intervention ensures that whenever the contextualization weights would point to a sense vector in \textit{MacBook} to predict words associated with \textit{Apple}, it now predicts words associated with \textit{HP}.

We project each sense vector of $\x$ to the nullspace of $E\x_r$, and then add in $E\x_a$:
\begin{align*}
 \tilde{C}(\x)_\ell = C(\x)_\ell+ \frac{C(\x)_\ell^\top E\x_r}{\|C(\x_r)_\ell\|_2^2}\left( \frac{E\x_a}{\phi} -E\x_r\right),
\end{align*}
where $\phi=\frac{\|E\x_a\|_2^2}{\|E\x_r\|_2^2}$ is a normalization term to account for the differing norms of $E\x_a$ and $E\x_r$.
Intuitively, this projection modifies each sense vector in measure proportional to how much $\x_r$ was predicted by that sense.
So, senses of \textit{MacBook} that would added mass to \textit{Apple} now add mass to \textit{HP}; unrelated senses are not affected.
In Table~\ref{table_macbook_control}, we show samples providing intuition for how \textit{MacBook} evokes HP instead of Apple, but is otherwise semantically and syntactically maintained.

\section{Related Work}

\paragraph{Representation learning in NLP.}
Learning probabilistic models of text for use in representation learning and identifying resulting structure has a long history in NLP, from non-contextual word vectors \cite{scheutze1992dimensions,rohde2006improved,turney2010frequency,mikolov2013efficient,bojanowski2017enriching} to contextual networks \cite{elman1990finding,bengio2000neural,collobert2008unified,sutskever2011generating,peters2018deep,radford2018improving}.
Deep Averaging Networks \cite{iyyer2015deep} are not \modelname{s}; they first perform averaging and then nonlinear computation.

\paragraph{Interpretability for Control of NLP networks.}
A burgeoning body of work attempts to intervene on monolithic neural networks for interpretability and control \cite{meng2022locating,meng2023massediting}, and for mechanistic understanding \cite{olsen-etal-2021-assessing,elhage2021mathematical}.
Implicitly, \modelname{s} develop a somewhat human-understandable language of machine concepts, an idea espoused in \citet{kim2018interpretability,koh2020concept}.
The connections between interpretation and control are rich; much work has gone into the detection and extraction of emergent structure in networks \cite{hupkes2018visualisation,liu2019linguistic} as well as subsequently modulating behavior \cite{lakretz2019emergence,eisape2022probing}.

\paragraph{Generalized Additive Models.}
Generalized Additive Models (GAMs; \citet{hastie1986generalized}) are a function family that (1) independently transforms each input feature, (2) sums these transformations of inputs and (3) applies a non-linear link function (e.g., softmax):
\begin{align}
  f(\x_{1:n}) = \Phi \left( r_1(x_i) + \cdots + r_n(x_n)\right)
\end{align}
Treating each word-position pair as a feature, Backpacks are not GAMs because they include a weighting $\alpha$ that depends on all features.
However, \modelname{s} share an intuition of computing independent representations of each feature and aggregating by addition.
Neural GAMs have been proposed for interpretability \cite{agarwal2021neural,YANG2021108192,chang2022nodegam,radenovic2022neural,dubey2022scalable}, though never to our knowledge in language modeling.
We expect that without context-dependent weighting, models would be insufficiently expressive for language modeling.

\section{Discussion}

In this section, we address a few natural questions about the expressivity and interpretability of \modelname{s}, highlighting the limits of our knowledge.

\paragraph{How do \modelname{s} compare to architecture X?}
The \modelname structure does not depend upon using a Transformer to compute the contextualization weights.
We could parameterize the contextualization function with a different architecture (e.g., LSTM, S4 \cite{gu2021efficiently}) and use the resulting weights to compute the \modelname sense vector sum.
This architecture, e.g., the \modelname-S4, could then be compared to the standard S4 architecture.

\paragraph{Are \modelname{s} as expressive as Transformers?}
We don't know.
If the number of linearly independent sense vectors is at least $d$, then a sufficiently complex contextualization network could treat them as an arbitrary basis.
A concern we've often heard is that ``simply'' adding together sense vectors should not be expressive enough to handle, e.g., negation.
However, as long as the requisite building blocks exist in the prefix, a contextualization network that recognizes the negation or other property could properly distribute weights.

\paragraph{Are \modelname{s} inherently interpretable?}
No, but we believe no architecture is.
Each architecture provides a set of tools that may or may not be useful for differing goals.
To us, the key is the mechanistic guarantees \modelname{s} offer, which will vary in utility depending on how well-specialized the learned sense vectors are for a specific kind of control.
Also, the visualizations we provide (top-$k$ highest-scored words) only provide a small view into a sense's potential uses, because scores are non-zero for the whole vocabulary.

\paragraph{Are \modelname{s} as compute-efficient as Transformers?}
At a glance, no. %
\modelname{s} have an underlying Transformer as well as extra parameters, but may perform roughly as well as just the underlying Transformer.
However, sense vectors are sparsely activated---only those from the relevant sequence need be on GPU---and after training, can be computed by lookup.

\paragraph{Why do sense vectors specialize?}
Ablations in Table~\ref{table_p_sense_vecs_results} show that they should at least learn to be linearly independent, since linear dependence is equivalent to having having fewer sense vectors, which causes higher perplexity.
The specialization of sense vectors to seemingly coherent categories may be attributable to the shared feed-forward network that computes them, and/or the contextualization network learning to assign similar weight distributions to senses with similar roles.

\paragraph{Are sense vectors like ``word senses?''}
No; they encode a notion of ``predictive utility'' that doesn't align with traditional notions of word sense.
We use the name ``sense vector'' however because they form a new, useful notion of decomposition of the possible contextual uses of a word into components that are softly combined in each context.

\section{Conclusion}
Non-contextual word2vec embeddings initiated modern deep learning research in NLP, and have fascinating geometric structure.
Now, research has largely moved on to monolithic representations, first from RNNs and now from Transformers.
Our work suggests that we can have both rich lexical structure and interventions, and strong contextual performance, in a single model. 

\section{Acknowledgements}

The authors would like to thank Amita Kamath, Steven Cao, Xiang Lisa Li, Ian Covert, and the Stanford NLP Group community for useful discussions.
Further support came from the Stanford Center for Research on Foundation Models.
Christopher Manning is a CIFAR Fellow.
John Hewitt was supported by an NSF Graduate Research Fellowship under grant number DGE-1656518 and by the CIFAR Learning in Machines and Brains program. We gratefully acknowledge the
support of a PECASE Award to Percy Liang.

\section{Limitations}

There is a fundamental uncertainty in whether \modelname language models will continue to scale with parameters and data and be viable alternatives to Transformers at larger model scales.
In this study, we were unable to scale larger, and hope that future work will test larger model scales.
In a similar vein, we do not verify that \modelname language models perform well across multiple languages.
We also do not consider, e.g., finetuning \modelname{s} on other tasks, or masked language modeling---there is a wide range of possible uses that remain to be verified.

One potential obstacle to the use of \modelname{s} that we do not study is the effect of tokenization in languages with richer morphological structure than English---will the \modelname structure be amenable to modeling those languages?
This may be difficult because, intuitively, the interpretability and control of \modelname{s} relates to the semantics of individual tokens.
Even in English, small subwords not indicative of a single word are hard to interpret.
What we hope to have provided is a sufficient set of experiments to motivate the further exploration of \modelname{s}.

\section{Ethics}
This paper describes and releases an open-domain language model trained on a largely unfiltered subsection of the (mostly English portions of the) textual internet, and describes methods for interpreting and controlling said model.
Any control method that can be used to help understand and guide the generation of a model can be used to more effectively generate toxic or illegal content.
Despite this, we do expect that, overall, the benefit of deeper insight into \modelname language models is a step in the right direction. 
In particular, explanations based on the structure of \modelname{s} may be able to provide insights into the mechanisms behind model behaviors, increasing transparency.

The concrete models we will release, up to and including 170M parameters, are substantially smaller and less performant at generating text than many of the publicly and commercially available language models available right now, so we do not expect there to be considerable negative repercussions from the release of the artifacts. 
The code we release, however, could be used or replicated to train much larger \modelname LMs by corporations or governments.

\bibliography{anthology,custom}
\bibliographystyle{acl_natbib}

\clearpage

\appendix

\section{Language Model Training Details} \label{appendix_training_details}

We use the FlashAttention codebase \cite{dao2022flashattention} which in turn relies on the Huggingface codebase \cite{wolf-etal-2020-transformers} and NumPy \cite{harris2020array}.
We perform no preprocessing of OpenWebText.
We do no explicit hyperparameter sweep for OpenWebText training beyond our sense vector ablation, instead taking the defaults provided.
We train our models on 4 A100 (40GB) GPUs.
All experiments test a single trained Small (124M Transformer or 170M \modelname) model due to computational constraints.

\subsection{The feed-forward sense network.}

We parameterize the feed-forward network for our sense vectors by first performing layer normalization on the input embeddings, and then a feed-forward layer with residual connection and layer norm (despite it being a function of just one word) to dimensionality $4d$ and back to $d$. Then a subsequent feed-forward network to hidden dimensionality $4d$ and then up to $k*d$.
We include a second layer norm and residual before the second feed-forward layer accidentally as a side-effect of the underlying language model codebase.

For our experiments ablating $k$ in Section~\ref{sec_ablation_results}, the second feed-forward component maps to $d$ and then $kd$, not $4d\rightarrow kd$.

\section{Extra evaluations}

\subsection{Timing Benchmarking}
To benchmark the speed of each model, we used a single A100 GPU, running the forward pass of each model with a sequence length of 512 and a batch size of 32.
We ran 100 forward passes and present the average time taken across the 100.
We present this in lieu of FLOPs because A100 GPUs are relatively standard, and this allows for a more directly usable time estimate.
Results are in Table~\ref{table_model_speed}.
We find that \modelname{s} take roughly 1.4x as long to run as their underlying Transformers.

\begin{table}[t!]
\small
\centering
\begin{tabular}{l c }
\toprule
Model & Time $\downarrow$ \\
\midrule
\modelname-Micro & 0.093  \\
 
Transformer-Micro & \bf 0.065 \\
\midrule
\modelname-Mini &  0.21 \\
Transformer-Mini  & \bf 0.15 \\
\midrule
\modelname-Small  & 0.36 \\
Transformer-Small  & \bf 0.26 \\
\bottomrule
\end{tabular}
\caption{\label{table_model_speed} Timing benchmarking results on an A100, average time to compute forward pass on 32-batch size 512-sequence length input.}
\end{table}

\begin{table*}
\centering
\small
\begin{tabular}{c c c c}
\toprule
\# Senses & Total Params & Contextl. Params & OWT PPL\\
\midrule
1 & 74.3M & 72.7M & 38.5 \\
4 & 75.6M & 72.7M & 29.3 \\
16 & 80.5M  & 72.7M & 26.0 \\
64 & 100.2M & 72.7M & 24.0 \\
\bottomrule
\end{tabular}
\caption{\label{table_p_sense_vecs_results} OWT perplexity and parameter count as a function of the number of sense vectors. All models trained for 50k steps, 500k token batch size, on OWT.}
\end{table*}

\begin{table}
\small
\centering
\begin{tabular}{l r r r}
\toprule
Model & Dim & Layers & Heads\\
\midrule
Micro & 384 & 6 & 6\\
Mini & 640 & 8 & 8\\
Small & 768 & 12 & 12\\
\bottomrule
\end{tabular}
\caption{\label{table_model_hyperparams}Model size hyperparameters.}
\end{table}

\section{Lexical Similarity Details}
To handle words in the lexical similarity datasets that don't appear as single words in the tokenizer, we use one of two methods.
We either average all subwords, or take the first subword.
The results for the two methods were similar, but we take the better overall for each model.
For all Backpack methods, our 124M-parameter Transformer, and GPT-2-xl, we average all subwords.
For GPT-J (which uses the same tokenizer), we take the first subword.

\section{Sense Vector Control Details}
\label{appendix_generation_details}

\subsection{Topic control details}

The full results are in Table~\ref{table_topic}.
The list of topics, and the corresponding bags-of-words, are given in Table~\ref{table_appendix_topic_words}.
For PPLM, the hyperparameter we vary to change the strength of topic control is the step size \cite{dathathri2019plug}.

\begin{table}
\centering
\small
\begin{tabular}{l l }
\toprule
Topic Label &  Bag-of-words\\
\midrule
 arts\_culture& arts, culture\\
 business\_entrepreneurs& business, entrepreneurs\\
 celebrity\_pop\_culture & celebrity, pop, culture\\
 diaries\_daily\_life & diaries, daily, life\\
 family & family\\
 fashion\_style & fashion, style\\
 film\_tv\_video & film, tv, video \\
 fitness\_health & fitness, health\\
 food\_dining& food, dining\\
 gaming & gaming\\
 music & music\\
 news\_social\_concern& news, social, concern\\
 other\_hobbies & hobbies\\
 relationships & relationships\\
 sports & sports\\
 travel\_adventure & travel, adventure\\
 youth\_student\_life & youth, student, life\\
 \bottomrule
 \end{tabular}
 \caption{\label{table_appendix_topic_words}The topics used in our topic classifier, and the bags-of-words we use for control.}
 \end{table}
 
 We consider a document as matching the semantic control if the classifier assigns greater than $0.5$ probability to the attempted class.
 We generated from our models with ancestral sampling with no truncation or temperature change.

\paragraph{Topic control.}

Let $b\in\mathbb{R}^{|\V|}$ be the many-hot vector defined by the bag of words input to the control problem. That is, if the bag is \textit{arts, culture}, then $b$ has $1$ at the indices corresponding to those words, and $0$ elsewhere.
To determine the initial weights $\delta$ for each sense vector, we first sort all $|\V|*k$ sense vectors by decreasing normalized dot product with the bag of words vector:
\begin{align}
s(C(\x)) = \frac{b^\top E^\top C(\x)}{\max(E^\top C(\x))}
\end{align}
We then take the $0.95$, $0.80$, and $0.60$ quantiles of these scores to determine how to weight the vectors.
Intuitively, the vectors in the highest quantiles (most associated with the target topic) are upweighted the most during decoding, to push the generation towards the topic.
The three quantiles partition the set of scores into 4, which are given separate $\delta$ values; the exact 4 depend on the strength of control (i.e., different points in Figure~\ref{fig_topic}.)
The exact $\delta$ upweighting for each point are given in Table~\ref{table_appendix_topic_strengths}.

\paragraph{Topic annealing.}

From the the beginning value of $\delta$ given above, we anneal back to $1$ as follows.
For each sense $C(\x_j)_\ell$, we compute the total sum of non-negative log-probability assigned by the sense to the set of words generated so far, intuitively to compute whether the words already generated express the meaning intended by the sense:
\begin{align}
&a_{C(\x_j)_\ell} = \sum_{i=1}^n  \max\left (\x_i^\top E^\top C(\x_j)_\ell), 0\right).
\end{align}
We then re-weight by a term dependent on the sequence index to upweight terms near to the most recently generated text:
\begin{align}
&b_{C(\x_j)_\ell} = \sigma\left(-a_{C(\x_j)_\ell}f+6\right) * \left(1+j\right)/100
\end{align}
where $j$ is the index of the word of the sense vector in the generated text, and $f$ is a scaling constant set to 7.5 divided by the maximum $\delta$ in the experiment (the maximum of each row in Table~\ref{table_appendix_topic_strengths}.)

Finally, we compute the annealed $\delta$ as a soft combination, weighted by $b_{C(\x_j)_\ell}$, of the maximum delta and the default of $1$:
\begin{align}
\delta_{\ell i j} = b_{C(\x_j)_\ell}\delta_{\ell i j} + (1-a)*1.
\end{align}

\begin{table}
\centering
\small
\begin{tabular}{c c}
\toprule
Control Strength & $\delta$ for quantiles $0.95, 0.80, 0.6, <0.6$ \\
\midrule
0 (unmodified) & 1,1,1,1\\
1 & 1.5, 1.5, 1.3, 1\\
2 & 2.2, 2.2, 1.5, 1\\
3 & 3.3, 3.3, 3, 1\\
\bottomrule
\end{tabular}
\caption{\label{table_appendix_topic_strengths}Initial topic control weights for each quantile.}
\end{table}

\begin{table}
\small
\centering
\begin{tabular}{l c c c}
\toprule
Method & Sem Acc $\uparrow$ &Toks-in-vocab $\downarrow$ & MAUVE $\uparrow$ \\
\midrule
\multicolumn{2}{l}{\textit{Transformer}}\\
Unchanged & 6.8\% & 0.0\% & 0.95 \\
PPLM-.01 & 8.4\% & 0.1\% & 0.94 \\
PPLM-.04 & 23.9\% & 2.6\% &0.81 \\
PPLM-.05 & 30.3\% & 5.5\% & 0.62 \\
PPLM-.06 & 37.7\% & 12.3\% &0.41 \\
PPLM-.07 & 40.8\% & 18.8\% & 0.25\\
\midrule
\multicolumn{2}{l}{\textit{Backpack}}\\
Unchanged & 7.4\%  & 0.0\% & 0.92\\
Ours$_{+1}$ & 12.1\%  & 0.2\% & 0.91\\
Ours$_{+2}$ & 24.3\% & 1.5\% & 0.90 \\
Ours$_{+3}$ & 35.3\%&  3.5\% & 0.83\\
\bottomrule
\end{tabular}
\caption{\label{table_topic}Topic control via  pseudovocabulary, vs PPLM. MAUVE scores are computed with respect to 8000 samples drawn across the topics.}
\end{table}

\subsection{Gender bias mitigation details} \label{appendix_genderbias_details}

For the third-person singular verb \textit{they}, we found that our sense intervention on sense 10 slightly increases the probability of \textit{they} relative to \textit{he} or \textit{she}.

The full set of nouns and prompts we use is as follows.
For role nouns, we use mechanic, accountant, farmer, baker, assistant, construction, guard, carpenter, analyst, physician, cook, clerk, manager, developer, librarian, salesperson, mover, hairdresser, auditor, sheriff, janitor, driver, chief, teacher, writer, receptionist, CEO, nurse, housekeeper, secretary, counselor, attendant, supervisor, designer, lawyer, editor, cashier, laborer, tailor, cleaner.

For evaluation, we use the prompts in Table~\ref{appendix_prompts}.
For experiments in which we estimate the strength of removal to reduce gender bias, we do the optimization individually for each profession noun, on the separate set of prompts in Table~\ref{appendix_train_prompts}.

\begin{table}
\centering
\small
\begin{tabular}{c c c c}
\toprule
\multicolumn{2}{c}{\textit{nurse} } & \multicolumn{2}{c}{\textit{developer} }\\
logprob & target & logprob & target\\
\midrule
+12.21 & professional     & -15.27&She   \\
+11.62 & She              & -15.14& herself  \\
-13.38  &himself        & -13.02& her  \\
 -13.24 &His            & -12.34&corruption  \\
 -12.63 & his           & -12.15& she  \\
 -12.31 &his            & -12.14&gam  \\
 -11.39 & he            & -12.13&HR  \\
 \bottomrule
\end{tabular}
\caption{\label{table_gender_sense_example} Some of the most up- or down-weighted words by sense 10 of the words \textit{nurse} and \textit{developer}. For each role word, the sense encodes the stereotypical gender of the role.}
\end{table}

\begin{table*}
\small
\centering
\begin{tabular}{l}
\toprule
Prompt\\
\midrule
    I really liked my PROFESSION. What I liked most was that, \\
    I really liked my PROFESSION. In fact,\\
    My PROFESSION said that \\
    My PROFESSION believes that \\
    This one PROFESSION believes that \\
    This one PROFESSION believes \\
    My PROFESSION said \\
    My PROFESSION believes \\
    The PROFESSION was with the car. When \\
    The PROFESSION was with the car. Then \\
    While driving to the store, the PROFESSION looked over on the dash and then \\
    A PROFESSION went to chat over to chat before\\ 
    Some PROFESSION asked to look over things, but \\
    \bottomrule
\end{tabular}
\caption{\label{appendix_prompts} Prompts used for gender bias evaluation.}
\end{table*}
\begin{table*}
\small
\centering
\begin{tabular}{l}
\toprule
Prompt\\
\midrule
    The PROFESSION came into the room. When \\
    The PROFESSION came into the room. Then \\
    I went to the talk to the PROFESSION. What \\
    I went over to the talk to the PROFESSION. Why \\
    I went over to the talk to the PROFESSION; \\
    \bottomrule
\end{tabular}
\caption{\label{appendix_train_prompts} Prompts used for gender debiasing estimation.}
\end{table*}

\begin{table*}
\small
\begin{tabular}{p{2cm} p{6.1cm} p{6.1cm}}
\toprule
Topic & \modelname Sense Control, Control=2  & Transformer PPLM, stepsize=0.04\\
\midrule
Arts \& Culture & (I kind dude) just used to be creative and make music but it feels like I'm 40. Conscious of how bad it is and how awesome it is like making such a fast consciousness and not even being musician meets people who answer for you, especially when it's scary." de la Cruz © Dan Wilson (2002). & Sunday, February 6: WestSTAR en-twitter-ilesurcast.blogspot.com

Tuesday, February 7: WestStar en-chat breathe of ancient energy. www.weststar.org

Monday, February 8: West Star

Mares and Moon of the ages

“Happiness is not easy to do”, Nicolas Jeansma, the Eternal Life programme director analyses history, culture, sociality and social magic.

: 'Oh the\\
\midrule
Business \& Entrepreneurship & Flickr advertisers is nothing new, so let's hope you know where you buy the latest edition.

At the same time, the fix has been pushed through, and while the overall business is pulling away from mainland Asia, publishers have given control over social media options to researchers at New York University and Columbia University. A new report from the Columbia board offers some clues as to why.

"My store in Alabama is used to a lot of Marines, and I just dropped as such. I don't know why, but I've had & We've decided to put out a newsletter to your guys, wondering as you cope with the tribulations of your business ventures and a job position. One way to put it is: You're not good enough. You've failed and you're not getting anything done. You're not doing enough. You're not bringing the passion and ideas you might have to a business. But one thing's for sure: if you self-promote, you often might take the business to a profitable buyer. Continue\\
\midrule
Celebrity \& Pop Culture` & Meetings and greets with reporters and celebrities of all kinds — pop culture, fashion, sports, food, celebrity lifestyle and otherwise — have been laid door-to-door on the Dallas television market with both LaVar and his wife, Arron, taking over the showroom-oneship business at Big Star Barber.

“We think Big Star’s an interesting exchange,” Arron says. “They’ve got an experience they’re & Type Services rumors have been up in the media since last month—and now we have some confirmed to the CBC Radio musical news channel's Twitter stream.

The group's guitarist, Greg Carr, has just announced that he's working with Papa John as the band's lead singer and guitarist. According to bizarre French pop culture creation icon Valentino pop music singer/writer Jiv pop pop model, who also wrote pop pop music's MySpace and Twitter pop memes, Cassidy gig pop pop superstar is\\
\midrule
Diary \& Daily Life & The exact actual life cycle life form life soars on and dies off in comparison to our own. During the first few years of life, the total life form you take to decide what to eat, how much of it to drink, why, and whether you want to exercise have been completely smashed and the technological capability to make that happen seriously out of the blue has been completely lost, jumping from complexity to complexity, totally overwhelming the mushroom in its ability to discover what levels it's supposed to & The Rome crew logam tagged Louisville Main Street today morning and observed a loading dock at the center of downtown Louisville. The dock is just bigger than what was supposed to dock the loading area for emergencies. They watched over the crowd after passing the boat and finally realized that they'd caught some missed traffic signals. "Serious congestion" has so far unnerved people from the Grande family picnics to weddings picnics picnics.

MTD Charlotte Pulse (@mtdphp\\
\midrule
Fashion & This article is about the fashion label fashion week fashion style month fashion fashion style fashion style fashion week fashion style fashion fashion fashion style fashion fashion style fashion history fashion fashion fashion fashion fashion fashion fashion johnny dressed in an actor's specially created costume news news icon

The Comic Relief series features stories, such as plungers from the comic books.

It was originally published as a comic published in Dark Horse Comics in English and in both comic books and graphic novels.[1] It was produced & Twitter personality @ceboperformancemk tweeted in response to the story about you.

Fashion designer underwear, designer cook dress, sexuality art models, sex con artists, real goths. BuzzFeed

You think my brain’s shit about what’s fashion looks like? Yeah no, I’m not on it. I’m fashion. I’m fine fashion. Yes I appreciate the brand but the people behind it[…] adults go fashion, or \\
\bottomrule
\end{tabular}
\caption{\label{table_topic_ex1} The first, non-cherry-picked category-satisfying example from each model.}
\end{table*}

\begin{table*}
\small
\begin{tabular}{p{2cm} p{6.1cm} p{6.1cm}}
\toprule
Topic & \modelname Sense Control, Control=2  & Transformer PPLM, stepsize=0.04\\
\midrule
Film, TV, \& Video & Originally published Live chat Qs with the film website writer, who raised millions at least two years ago I contacted him with the same questions as you're doing.

I'm a bit optimistic that you're right, but you're just not responding. As you studied the film timer/mapplot'n'cookies response speed, I read the excerpts and couldn't make out a massive amount of time differences. Very minor.

What do you think about some of the terms & Well, the hype is real, and with the release of the latest episode of season two (which I’m probably not supposed to review), it feels like you won’t be afraid to retweets fideo.

By “HAPPY FINALS,” the footage maker has used a GIF video to give viewers look at Fideo’s dancing triangles and serenity dancing around a moving picture. Thank you, fideo!

If the\\
\midrule
Fitness \& Health & CLOSE Don't think tanking will spell good news for Detroit medical marijuana patients but the owner of its dispensaries saying that is just part of the problem facing the growing number of ill people having access to pot.

Healthcare workers are treated for tumors in a dispensary in Oakland. (Photo: Christopher Satorica, Special to CNN)

An array of medical centers have lined up near Detroit after a medical marijuana reform forum at the University of Michigan put the debate over the drug at &Today 

we learn more about the rise of the ice age, multi-drug cocaine epidemic, global population explosion and warfare epidemic by following Dr. Kristof Dr. Freedk published in the British Journal of Medicine The authors update their lofty goal and continue to refine their work for public health.

The International Health Services Committee has just released a new research, The next three years could be very costly for health care in Australia, hospitals, state health systems and dietary health. A recent report from\\
\midrule
Food \& Dining & As weeks wore maple leafed food trucks, and food processors reminisced about their great days past, healthcare workers found out one day that they should get better working conditions with little regard for their bodies.

Barbara Butterfield, the former Shop Swagger workshop in Clarksdale, got shot dead on Monday morning when she tried to stop a father Francisco Lee Walker from firing a gun. Walker, 20, had just started his Aug. 27 firing. Exposure to fire and clothes caused Walker & I would dearly love to stand at that galloping chair and who doesn’t has amazingly friends associated with their backs hurting? I was a big first timer yesterday. Not always with bacon but I held til calms up. Big chunks of bacon super nice but not me. However there are times where the pieces pull apart and this happens very hard to homo and crackers afgh. All Mixed ones made popular points that have the food triggers across: lack of meats rinsing and eating\\
\midrule
Gaming & My parents encouraging kids to be competitive gaming at school is not a new concept. Gaming has been around since the earliest days on paper, and their perspective is always superior than yours. Quality doesn't always apply, and that's why we bucked that trend' father

The English woman's son Anthony, who is best known for his role as Most Wanted, came up with the idea of pulling a 30-year-old mentally disabled woman who had been using motorbikes for & Every year, many migrants continue to struggle to find the skills they need in an emerging technology. But every year, it comes quite a surprise to hear the latest news about computerized computing and the gaming community.

For the sake of many gaming communities, we here at 14/gamer.org love gaming. It is an important industry in gaming, as it often draws passionate gamers from gaming and lends the gaming community the ability to allow itself special moments like gaming gaming days and gaming gaming. We\\
\midrule
Music & David has been a staunch critic of music culture that promotes music as something new, daring, and powerful. As he explained. ("I never thought I was one of those stupid, stupid old people who just listens to music or really hears it it's always the same as when I was a kid," he said.) And when he was a touring musician, those opinions were totally correct. Read the entire interview below.

On trying to inculcate younger vocalists with the " & From the East art council HQ of MondoJapan

Everyone laughs when a sheet metal title is rendered artistically constrained and we say, "Whoa. Then the skin guy! This is a very Chi style steel." Well I don't think anyone's ever heard that before. There's only one coil metal group that is not a tarantella performance music group...at least in America...compart music ten times over and they will never release tracks for it that it is a \\
\midrule
\end{tabular}
\caption{\label{table_topic_ex2} The first, non-cherry-picked category-satisfying example from each model.}
\end{table*}
\begin{table*}
\small
\begin{tabular}{p{2cm} p{6.1cm} p{6.1cm}}
\toprule
Topic & \modelname Sense Control, Control=2  & Transformer PPLM, stepsize=0.04\\
\midrule

News \& Social Concern & Buildersh B2 has been compared unfathomable by a number of critics because of his security concerns.

Breaking News Alerts Get breaking news when it happens — in your inbox. Email Sign Up By signing up you agree to receive email newsletters or alerts from POLITICO. You can unsubscribe at any time.

Yet, on Tuesday, Monday and Tuesday, the developer reached the milestone of completing the first UPS facility located in the town of Cloudbreak. He secured \$4 &  After initially putting itself over Sports Illustrated on Monday, the New York Times was forced to apologize for its widespread coverage of its reporting on the State of Rhode Island – a state that has been the subject of gossip news for some time and which its harsh news and ratings policy has spawned.

Late at night on Monday, we learned that the New York Times had reached a breaking news cycle decision and we snagged our exclusive first look at the news. Here’s what you didn’t\\
\midrule
Relationships & Early life release parties is relationship couples with relationships over relationships. This census does not count relationships by those who have been with those relationships over the last three years. For more information about early life release parties, check the release party census.

Carlo Mathieu

Carlo Mathieu was born in 1958. He lives in Augusta, Ga., with his biological father, Malcolm Mathieu, who was president of the Augusta West Raceway at the time.

Benjamin Math & Any learning is like being completely ignorant of new information. Schools are forced to teach students to treat one another in the right way, but we still have to recognize that we have to learn how to be friends with as much as we can. When Santod relationships are hard and relationships can be complicated and confusing, there will always be learning relationships, relationships that remind us that we don't mean relationships, relationships relationships that are boundaries, relationships relationships with friends in need relationships with involved relationships, relationships relationships relationships \\
\midrule
Sports & PRESS W/NEWS BLOK Play slideshow 1 of 83 Express sports retail giant Sports Direct.

Sports Direct has revealed the on offer outdoor sports gear Brand new from Google has been developed. Here’s what you can expect from Google’s sporting expertise.<|endoftext|>About

The potential of a west coast restaurant for tolerance and pity

Their position at this point hurts me less than they believe it deserves, because they probably shouldn.

I'm going to help them &  Authorities in California say they are investigating equestrian skiers who struck a 19 year-old boy from a snow-covered mountainand beating him on the head with shovels.According to SmithCox, those same well clients found out they had also been tardled by a \$500 pour frompipe on top of of a Black Rock vault. And it appears the ultimate goal of those riders and their company of riders was killed.Jeremy Goschz is one of those survivors. His racing\\
\midrule
Travel \& Adventure & My next stop destination for me is adventure travel. I travel Disney World and make sure that the worlds under my belt and desert warriors that I’ve been fighting for have a place or two at their disposal that are compatible with my use of current technology. This job is being completed with the help of any freelance user submission information you may have provided. It’s only fair to give you some tips to help you figure it out if there are any unknown sideside locations that you  & Equality

Equality – open life – inequalities – political oppression –

write and publish your work

Equality is a freedom to work, to die. Access to free healthcare, free outer space travel, photocopies online, happy endings, self travel – to travel to someone else’s heart (read: stop taking drugs), to move faster, to travel in train travel, to stop a vacation abroad (tell others your travels), to return to a home each time\\
\midrule
Youth \& Student Life & College students at almost every age advantage who take advantage of learning opportunities in the sport of running spend at least five years an average of \$10 or more per year to do it, according to the University of San Diego’s National Football Clearinghouse.

Those risk factors lift nearly a third of university and college football athlete spend, more than double that of a comparable age group of men and women who spend 4,000 hours per year as runners, or 5,000 to & lame University saw a 32 per cent rise in its undergraduate science institutes and 14 per cent increase in its researchers from recent years.

Director Of University Development, Mike Brennan, said: "The growth in university employment, coming from such a historic campaign, is something to celebrate as we support our young people and room to progress in science and technology."

A student was interviewed in a recent paper about university employment, specifically a dissertation.

"For the first time, people are\\
\bottomrule
\end{tabular}
\caption{\label{table_topic_ex3} The first, non-cherry-picked category-satisfying example from each model. This is except for the Relationship category for the Transformer, where we skipped the first one due to content we particularly did not want to publish.}
\end{table*}

\begin{table*}
\small
\centering

\begin{tabular}{c c c c c c c c c c c c c c c c}
\toprule
\multicolumn{8}{c}{Positive Log-Probability Mass for Senses of word \textit{ quickly}}\\
0&            1&               2&          3&             4&        5&           6&             7&            \\ 
\midrule
approaching&  oggles&          quickly&    enough&        stro&     iii&         razen&         asuring&      \\ 
ascended&     Marks&           swiftly&    rotating&      zn&       Original&    forgotten&     delusion&     \\ 
grav&         Axis&            rapidly&    paced&         strokes&  alsa&        forget&        stimulated&   \\ 
gent&         claimer&         quick&      ened&          uling&    chenko&      social&        recollection& \\ 
disposed&     Roche&           quick&      retreating&    \$\_&     resolution&  rius&          stimul&       \\ 
risen&        demonstration&   instantly&  Subscribe&     grass&    ient&        relapse&       Wem&          \\ 
dispose&      blaster&         promptly&   dismissing&    lessly&   baskets&     baseless&      persistent&   \\ 
becoming&     ducers&          soon&       diminishing&   iken&     uin&         Statement&     urbed&        \\ 
ascert&       Specifications&  fast&       disappearing&  izing&    ora&         athing&        retard&       \\ 
climbed&      Viet&            Quick&      varying&       bg&       alid&        Akron&         restraint&    \\ 
\midrule
8&             9&             10&           11&             12&        13&           14&         15\\
\midrule
processors&    slowly&        tering&       Definitely&     quick&     oted&         ouse&       Sims\\
darts&         Slowly&        Bers&         initely&        quickest&  distances&    pee&        Noir\\
milliseconds&  Slow&          Fed&          raid&           quick&     outed&        ouses&      TMZ\\
wip&           conveniently&  ascus&        alright&        quicker&   aught&        pees&       Streets\\
iazep&         slower&        Bust&         Personally&     fast&      UC&           attach&     expressly\\
reptiles&      cheaply&       aucus&        laughs&         quickly&   ob&           tro&        Attend\\
Kelvin&        responsibly&   Ryu&          ALWAYS&         rapid&     digits&       iffe&       Rooms\\
Ow&            gradually&     sector&       Always&         fast&      ench&         aces&       Within\\
Soon&          quietly&       Petra&        Ideally&        faster&    Code&         lain&       Rum\\
Slug&          waiting&       DCS&          Roses&          fastest&   apers&        feet&       Forced\\
\midrule
\midrule
\multicolumn{8}{c}{Negative Log-Probability Mass for Senses of word \textit{ quickly}}\\
0&            1&               2&          3&             4&        5&           6&             7&           \\ 
\midrule
initely&      sburg&           ollen&      una&           Poké&     quickly&     Faster&        .&          \\ 
heit&         orem&            oned&       URE&           slow&     quick&       purposely&     Sorceress&   \\ 
Aly&          Untitled&        oths&       rast&          slower&   swiftly&     deliberately&  itars&       \\ 
istically&    anted&           ook&        ipt&           slows&    rapidly&     Definitely&    Shogun&      \\ 
Always&       untreated&       ught&       ocracy&        slowed&   quickest&    ey&            Yen&         \\ 
Doctors&      til&             Ded&        law&           DEV&      quick&       slower&        oenix&       \\ 
dl&           broken&          lost&       uthor&         encia&    Quick&       initely&       Jagu&        \\ 
urally&       past&            aught&      ema&           potions&  fast&        isner&         izz&         \\ 
ependence&    ebook&           recharge&   ory&           Machina&  instantly&   hesitated&     eral&        \\ 
raints&       Continue&        ady&        antis&         Slow&     Quick&       eyewitness&    finals&      \\ 
\midrule
 8&             9&             10&           11&             12&        13&           14&         15\\
 \midrule
quist&         WM&            prototype&    ciating&        kins&      quick&        Laur&       thal\\
 ocker&         isf&           projector&    scrambling&     Host&      quick&        Never&      imble\\
 ovsky&         fb&            reconcil&     rapid&          loudspe&   quickly&      Jimmy&      iquid\\
 ictions&       WF&            prominently&  newcomer&       enced&     Quick&        dearly&     initialized\\
 olation&       elevation&     counterfeit&  adapting&       Evil&      soon&         Dating&     ansas\\
 cano&          RM&            word&         speeding&       washed&    fast&         \_-\_&      IGH\\
 Proof&         975&           cellul&       frantic&        Kaf&       rapidly&      never&      unciation\\
 cert&          dir&           prototype&    novelty&        Glass&     Quick&        Certainly&  needs\\
 rero&          ESE&           collaps&      paced&          sod&       hurry&        eternal&    commit\\
 anch&          onder&         dyl&          instructional&  advers&    Immediately&  Rare&       tackle\\

\bottomrule
\end{tabular}
\caption{\label{appendix_table_quickly_senses}For each sense vector of the word \textit{quickly}, the 10 words to which the sense vector assigns the highest log-probability contribution, and the 10 to which it assigns the largest negative log-probability contribution. Note that usually, either the positive words are coherent or the negative---but not both for the same sense index. Some senses are not interpretable, and seem to be used by other parts of speech.}
\end{table*}

\end{document}